\documentclass[10pt,twocolumn,letterpaper]{article}

\usepackage{times}
\usepackage{epsfig}
\usepackage{graphicx}
\usepackage{amsmath}
\usepackage{amssymb}
\usepackage{diagbox}

\date{}

\begin{document}

\title{AutoGAN: Robust Classifier Against Adversarial Attacks}

\author{Blerta Lindqvist\\
Rutgers University\\
{\tt\small b.lindqvist@rutgers.edu}
\and
Shridatt Sugrim, Rauf Izmailov\\
Perspecta Labs\\
{\tt\small \{ssugrim, rizmailov\}@perspectalabs.com}}

\maketitle

\begin{abstract}
Classifiers fail to classify correctly input images that have been purposefully and imperceptibly perturbed to cause misclassification. This susceptability has been shown to be consistent across classifiers, regardless of their type, architecture or parameters. Common defenses against adversarial attacks modify the classifer boundary by training on additional adversarial examples created in various ways. In this paper, we introduce AutoGAN, which counters adversarial attacks by enhancing the lower-dimensional manifold defined by the training data and by projecting perturbed data points onto it. AutoGAN mitigates the need for knowing the attack type and magnitude as well as the need for having adversarial samples of the attack. Our approach uses a Generative Adversarial Network (GAN) with an autoencoder generator and a discriminator that also serves as a classifier. We test AutoGAN against adversarial samples generated with state-of-the-art Fast Gradient Sign Method (FGSM) as well as samples generated with random Gaussian noise, both using the MNIST dataset. For different magnitudes of perturbation in training and testing, AutoGAN can surpass the accuracy of FGSM method by up to 25\% points on samples perturbed using FGSM. Without an augmented training dataset, AutoGAN achieves an accuracy of 89\% compared to 1\% achieved by FGSM method on FGSM testing adversarial samples.
\end{abstract}

\section{Introduction}

\begin{figure}[t!]
\begin{center}
   \includegraphics[width=1.0\linewidth]{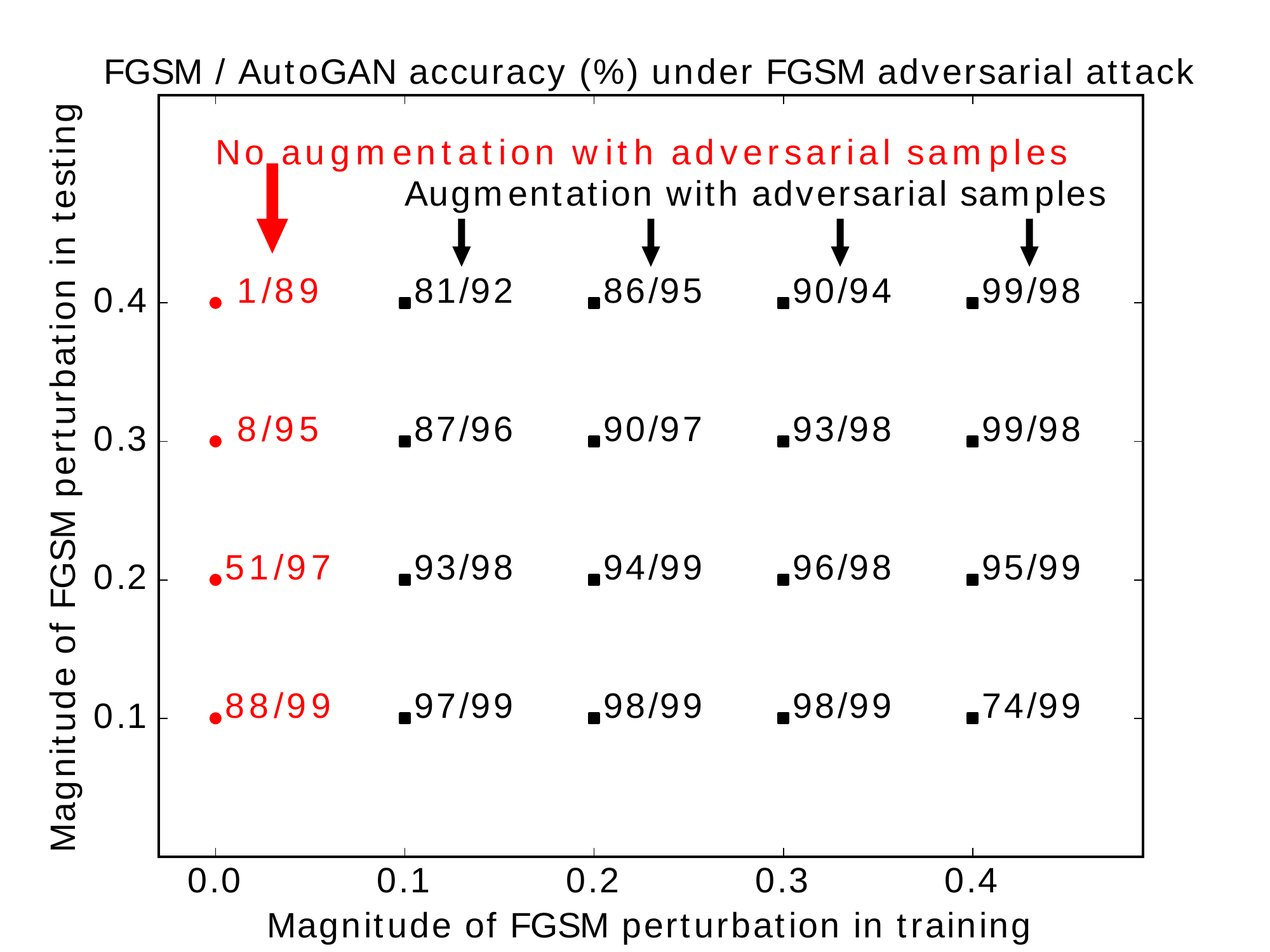}
\end{center}
   \caption{We show FGSM and AutoGAN method accuracy results for different magnitudes of FGSM perturbation in training and testing. The x axis displays the magnitude of FGSM perturbation in training, and the y axis displays the magnitude of FGSM in testing. Where the training dataset is not augmented with additional adversarial samples, AutoGAN accuracy surpasses FGSM method accuracy the most -- in one case 89\% to 1\%. Where the training dataset is augmented with adversarial samples, AutoGAN maintains high accuracy. FGSM method accuracy deteriorates when the magnitudes of perturbation in training and testing do not match -- bottom right and top left of the chart. AutoGAN values are means of three measurements, FGSM values are measured once due to no variance.}
\label{fig:fig0}
\end{figure}

Neural network classifiers misclassify adversarial examples~\cite{szegedy2013intriguing, biggio2013evasion}, which are original input samples with imperceptible perturbations. There are two attributes of these attacks~\cite{goodfellow6572explaining, papernot2016transferability, tramer2017space}: imperceptibility and transferability. Imperceptability (for visual datasets) means that these adversarial examples are largely indistinguishable from the original images to the human eye. Transferability~\cite{papernot2016transferability} means that perturbed images cause misclassification across classifiers of various types, structure and parameters~\cite{szegedy2013intriguing}. The increasing deployment of Neural Networks (NNs) in security and safety-critical domains such as autonomous driving~\cite{amodei2016aisafety}, healthcare~\cite{faust2018deep}, and malware detection~\cite{cui2018detection} makes countering these attacks important.

One common defense against adversarial attacks on machine learning applications is adversarial training~\cite{kurakin2016adversarial}, which often assumes advance knowledge of the type and magnitude of adversarial attack and availability of adversarial samples of that attack. These are strong assumptions that cannot easily be satisfied in machine learning applications. Even if the attack type were known, adversarial samples for a particular attack might not be easy to obtain. Hence, the need for generic algorithms against adversarial attacks that can handle various magnitudes of perturbation even without having additional adversarial samples to train on.

\begin{figure}[t!]
\begin{center}
   \includegraphics[width=1.0\linewidth]{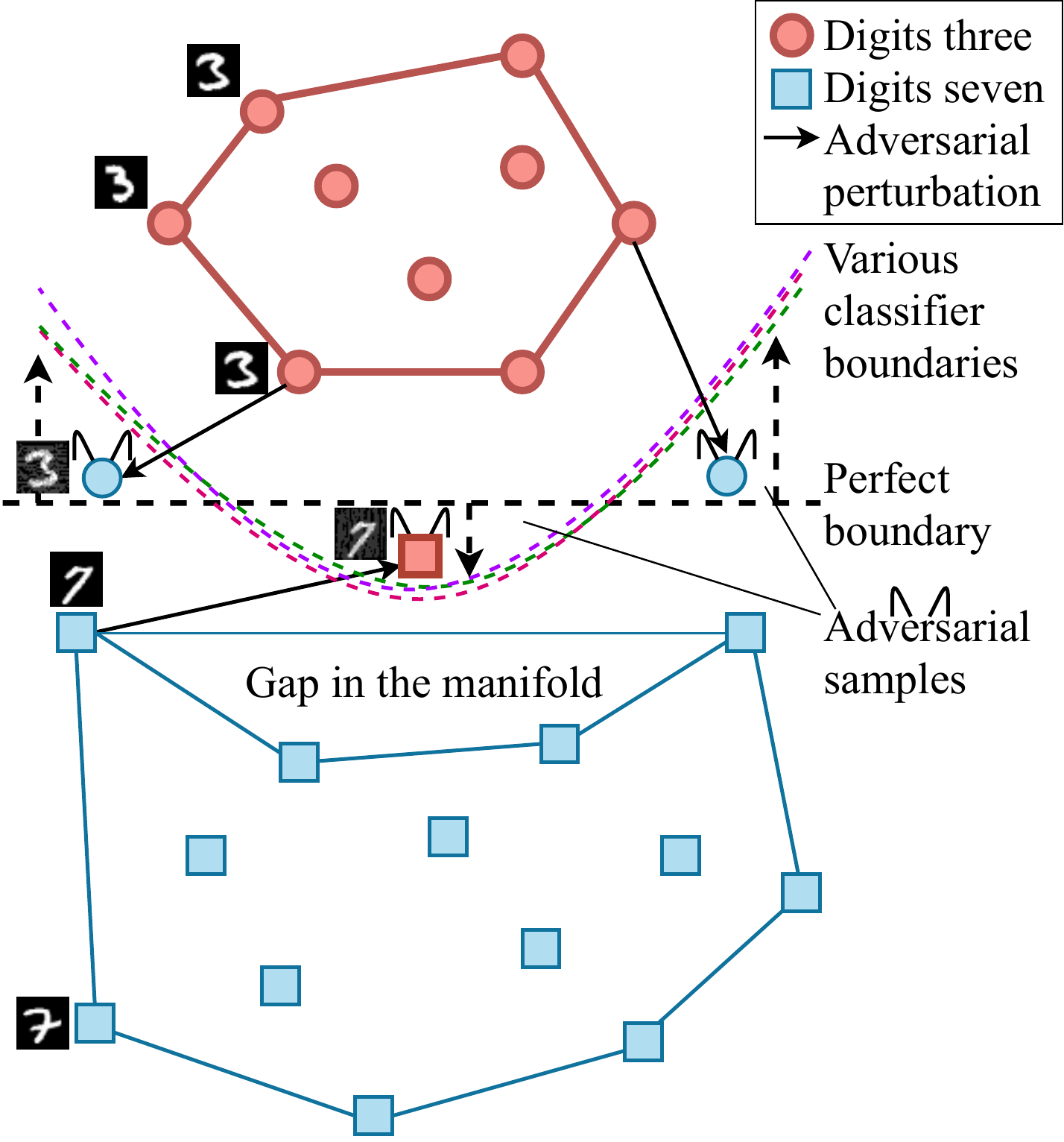}
\end{center}
   \caption{Illustration of the manifold gaps that cause adversarial attacks. There is a small pocket of space in the manifold that lacks data points from the square class. Classifiers aim to maintain same distance from points of different classes. This causes them to pass roughly in the same region of space. Due to the manifold gap, all classifiers shift towards the gap to maintain same distance from points of different classes. As a result, in the vicinity of the gap, all classifiers are surrounded by space the points of which would belong to one of the classes. They all misclassify the data points shown as adversarial samples, which are between the classifiers and the perfect boundary.}
\label{fig:fig6}
\end{figure}

In this paper, we design, implement and evaluate AutoGAN, a classifier against adversarial attacks that does not necessitate augmentation of the dataset with adversarial samples. Instead, we enhance the manifold of natural images and project the data points onto it. It is commonly accepted that the data points of natural image datasets occupy low-dimensional manifolds in the high-dimensional input space~\cite{Goodfellow-et-al-2016}. Manifolds are defined as collections of points in the input space that are connected. Each class has a manifold associated with it, and data points are classified by considering the proximity of the point to each class manifold, as illustrated in Figure~\ref{fig:fig6}. A trivial example of a manifold would be a letter S drawn on a two-dimensional paper surface. The space is two-dimensional (the paper it's drawn on), but the letter S (the line) is one-dimensional because if pulled from both ends, it would occupy only one dimension. AutoGAN uses a GAN with an autoencoder generator and a discriminator that also acts as the classifier. The presence of class labels in the classifying discriminator makes AutoGAN reshape the manifold for better classification. The major contributions of this paper are:

\begin{itemize}
  \item AutoGAN overcomes the absence of adversarial trainings samples, which is important in cases where attack type is not known and attack samples are not available. Figure~\ref{fig:fig0} shows that without an augmented dataset, AutoGAN gets an accuracy of 89\% compared to 1\% achieved with FGSM method on FGSM adversarial samples.
  \item AutoGAN is able to handle different magnitudes of training/testing adversarial perturbation, which is relevant when the magnitude of an attack is not known in advance. Figure~\ref{fig:fig0} shows that AutoGAN can surpass the accuracy of FGSM method by up to 25\% points when the training dataset is augmented.
  \item AutoGAN handles different kinds of adversarial attacks: samples from adversarial directions generated with state-of-the-art FGSM method -- Section~\ref{fgsmExp}, and samples from random directions generated with Gaussian noise per pixel -- Section~\ref{gaussExp}.
  \item AutoGAN is less exposed to adversarial attacks since the logic of countering these attacks is contained in both the generator and the discriminator, Section~\ref{autoganManifold}, but only the discriminator can be accessed by attackers.
\end{itemize}

\section{Related work} \label{relatedSection}

Vulnerability to adversarial attacks has been attributed to several factors. Among these factors are the extreme nonlinearity of deep networks~\cite{szegedy2013intriguing}; and linear behavior in high-dimensional spaces~\cite{goodfellow6572explaining}, leading to the FGSM method of adversarial direction. Szegedy \textit{et al.}~\cite{szegedy2013intriguing}, Song \textit{et al.}~\cite{song2017pixeldefend} and Tram{\`e}r \textit{et al.}~\cite{tramer2017space} have attributed the susceptibility to adversarial attacks  to low probability pockets in the manifold.

\paragraph{GAN framework} First introduced by Goodfellow \textit{et al.}~\cite{goodfellow2014generative}, the GAN framework sets two networks, a generator and a discriminator, against each other with competing losses. The generator has the goal to find a function that maps a random input vector to a realistic image. The discriminator has the goal to distinguish the generated image from real images. Among improvements to the GAN framework, the generated images have been made more realistic by making the generator network an autoencoder~\cite{zhu2016generative}, where the generator input is not a random vector but a dataset image. This enables generated images to stay close to the manifold of natural images~\cite{zhu2016generative}. An autoencoder generator also enables the latent representation of the autoencoder to represent the low-dimensional manifold of the input data and allows for navigation of the manifold, for example, interpolation between two points along the manifold. Another improvement to the GAN framework has been the incorporation of the class labels into the discriminator labels, for example, for better image quality~\cite{salimans2016improved, odena2016conditional} or for label efficiency~\cite{odena2016semi}.

\paragraph{Generation of adversarial samples} Many efforts have explored generation of adversarial samples. The FGSM method~\cite{goodfellow6572explaining} identifies directions in the input space that reduce the most the probability of correct classification as adversarial directions. Given a target model $h$, an input image $x$, its label $y$, the loss function $L(h(x),y_{true})$, FGSM creates adversarial samples by adding perturbation to input images:

\begin{equation}
x_{FGSM}^{adv}=x+\epsilon  \cdot sign( \triangledown_{x}L(h(x),y_{true})).
\end{equation}

In addition to FGSM adversarial directions, random adversarial directions can also be identified, for example with random Gaussian noise added to pixels~\cite{szegedy2013intriguing}. Sabour \textit{et al.}~\cite{sabour2015adversarial} generate adversarial samples by manipulating the deep representation of images to make them look like other specific images to neural networks. Ensemble Adversarial Training~\cite{tramer2017ensemble} augment the training data with perturbations from other models. Ensemble Adversarial Training decouples the generation of adversarial samples and the model being trained to counter adversarial attacks because adversarial perturbations can be transferred from one NN to another due to the transferability of adversarial attacks. Other efforts use GANs~\cite{goodfellow2014generative, radford2015unsupervised} to generate adversarial training examples~\cite{xiao2018generating, baluja2017adversarial, hu2017generating, kos2018adversarial}. Xiao \textit{et al.}~\cite{xiao2018generating} generate adversarial examples with adversarial networks, where GANs generate the perturbations. Kos \textit{et al.}~\cite{kos2018adversarial} utilize variational autoencoder (VAE) and VAE-GAN for generation of adversarial training examples. Baluja \textit{et al.}~\cite{baluja2017adversarial} uses two approaches for generation of adversarial samples: perturbation in residual blocks in residual networks; adversarial autoencoding using regularizers.

\paragraph{Defense against adversarial attacks} Current defense approaches against adversarial attacks can largely be classified into two types. The first type are methods that make the model architecture more robust. The second type are methods that change the adversarial samples to make them less adversarial.

The most common method of the first type is adversarial training by Kurakin \textit{et al.}~\cite{kurakin2016adversarial}, where the dataset is augmented with adversarial samples, often of the same kind as the attack, and the classifier is retrained. Adversarial training augments the dataset by filling out low-probability gaps with additional data points and then retraining the classifier to find a better boundary. The drawback of adversarial training is that it only provides defense against the attack model that was used to generate the augmentation adversarial samples and not against others. FGSM~\cite{goodfellow6572explaining} method uses adversarial training. Virtual Adversarial Training (VAT) by Miyato \textit{et al.}~\cite{miyato2017virtual} identifies directions that can deviate the inferred output distribution the most and smoothes the output distribution in these anisotropic directions. VAT is also applicable to semi-supervised applications. Defensive distillation~\cite{papernot2016distillation} by Papernot \textit{et al.} extracts knowledge for improving a network's own resilience to adversarial attacks. It is based on the distillation training procedure~\cite{hinton2014distilling}, which trains a NN using knowledge from a different NN. Defensive distillation adapts distillation to use the knowledge to improve its own resilience, instead of transferring it to another NN. In addition, it makes generation of adversarial samples more costly by increasing the average minimum number of features that need to be changed for generation of adversarial samples. Madry \textit{et al.}~\cite{madry2017towards} use robust optimization towards deep learning models resistant to adversarial attacks. Izmailov \textit{et al.}~\cite{rauf2018james} identify enablers of adversarial attacks and focus on featurization of data as a method to innoculate against adversarial examples.

Second type methods identify adversarial samples and move them towards the data distribution or the natural data manifold. Defense-GAN by Samangouei \textit{et al.}~\cite{samangouei2018defense} uses a GAN with a generator to project adversarial points onto the manifold of natural images. Given an input point that is potentially adversarial, Defense-GAN does several random initializations in the latent space and chooses from them a latent space seed that generates the closest match to the input. The closest match is considered as the projection of the original input point, though due to the randomness and depending on the number of random initializations, this projection might end up far from the real projection of the point on the manifold. Defense-GAN uses a classifier to which GAN input or GAN output or both GAN input and GAN output data points are used. Magnet by Meng \textit{et al.}~\cite{meng2017magnet} also moves adversarial samples towards the manifold of natural images by utilizing an autoencoder or a collection of autoencoders. It contains several detector networks that detect adversarial examples from the distance between the original image and the reconstructed image. The reformer network moves adversarial samples towards the manifold. Jha \textit{et al.}~\cite{milcom2018manifold} focus on identification of adversarial samples. They first identify a low-dimensional manifold where training samples lie. Then, based on the distance of new observed data points from this manifold, points are identified as adversarial or not. PixelDefend by Song \textit{et al.}~\cite{song2017pixeldefend} proposes generative models to move adversarial images towards the distribution seen in the data. PixelDefend identifies adversarial samples with statistical methods ($p$-value) and finds more probable samples by generating similar images and looking for highest probability images within a distance from a given image.

\paragraph{Summary} Existing methods either change classifiers to make them more robust to adversarial attacks or move adversarial points to make them less adversarial, but not both because it is not straightforward to do so. In contrast, AutoGAN employs both of these defenses by enhancing the manifold and changing its discriminator classifier at the same time as moving adversarial points toward the enhanced manifold to make them less adversarial. In contrast to other methods, AutoGAN does not identify adversarial samples explicitly.

AutoGAN uses the GAN framework because it has been successful in reconstructing manifolds of natural images. In distinction to other methods that also use GANs against adversarial attacks, the generator of AutoGAN is an autoencoder because it outputs more realistic images and enables navigation of and better adherence to the manifold. In the discriminator, AutoGAN incorporates class labels into both categories of discriminator labels, TRUE and FALSE which results in the AutoGAN discriminator having twice the number of labels of the dataset. Other methods that are doing the same only incorporate labels into one category. AutoGAN uses its discriminator also as a classifier, which is not only parsimonious for resources, but also utilizes what the discriminator has already learned.

We evaluate AutoGAN against adversarial samples in adversarial directions generated with the FGSM method~\cite{goodfellow6572explaining}. To further assess that the scope of AutoGAN is not only limited to adversarial directions, we evaluate AutoGAN against adversarial attacks in random directions created by adding random Gaussian noise to each pixel.

\section{Our approach: AutoGAN}

In this section, we detail our approach for countering adversarial attacks. AutoGAN belongs to both types of common approaches against adversarial attacks described in Section~\ref{relatedSection}. It belongs to the first type because AutoGAN changes the manifold of images by enhancing it. It also belongs to the second type of approaches because AutoGAN moves adversarial samples towards the manifold by projecting them onto it. By design, our AutoGAN method is agnostic of attack method and does not need additional adversarial training points, but can benefit from them.

\paragraph{Motivation} In our view, adversarial attacks can be explained with the presence of gaps in the manifolds of natural images~\cite{szegedy2013intriguing, song2017pixeldefend, tramer2017space}, caused by non-uniformly distributed sampling of the data points. These gaps cause widening of the margin between the points of different classes. Classifiers, in general, aim to maintain the same distance from points of different classes. Since the manifold gap affects the in-between space between classes, the boundaries of the classifiers are affected as well. The boundaries get shifted towards the gap region, as depicted in Figure~\ref{fig:fig6}.

Both defining characteristics of adversarial attacks are explainable with this view that explains adversarial attacks with gaps in the manifold of images. The transferability of adversarial attacks across diverse classifiers can be explained because though they may be diverse, classifiers will aim to place boundaries with roughly equal margins from the data points of the different classes, as illustrated in Figure~\ref{fig:fig6}. Having parts of the manifold unpopulated changes the margins between the classes and causes the classifying boundary to shift towards the area of the manifold where the gap is. Here, in the vicinity of the shifted boundary, the space on both its sides belongs to the same class, yet only one side is classified correctly. As can be seen in Figure~\ref{fig:fig6}, crossing the various classifier boundaries in the area close to the gap causes misclassification, however small the changes may be. This would explain the imperceptibility property of adversarial attacks, their second property. 

The manifold gap view also explains what the commonly employed defense -- adversarial training~\cite{kurakin2016adversarial} -- does. Adversarial training fills up the unpopulated gaps in the manifold with additional points, after which retraining on the augmented dataset redraws the classifying boundary. However, AutoGAN aims to cause the classifying boundary to be redrawn without augmenting the original dataset with additional adversarial samples by repositioning the points for better classification, thus changing the shape of the manifold towards better classification. This is motivated by the lack of knowledge of the attack type and magnitude, and the potential unavailability of adversarial samples for the particular attack. This knowledge and these samples are assumed to be available in various methods against adversarial attacks. Though AutoGAN does not necessitate augmenting the training data with adversarial points, it can benefit from additional adversarial points, as shown in Figure~\ref{fig:fig0}.

\begin{figure}[t!]
\begin{center}
   \includegraphics[width=0.8\linewidth]{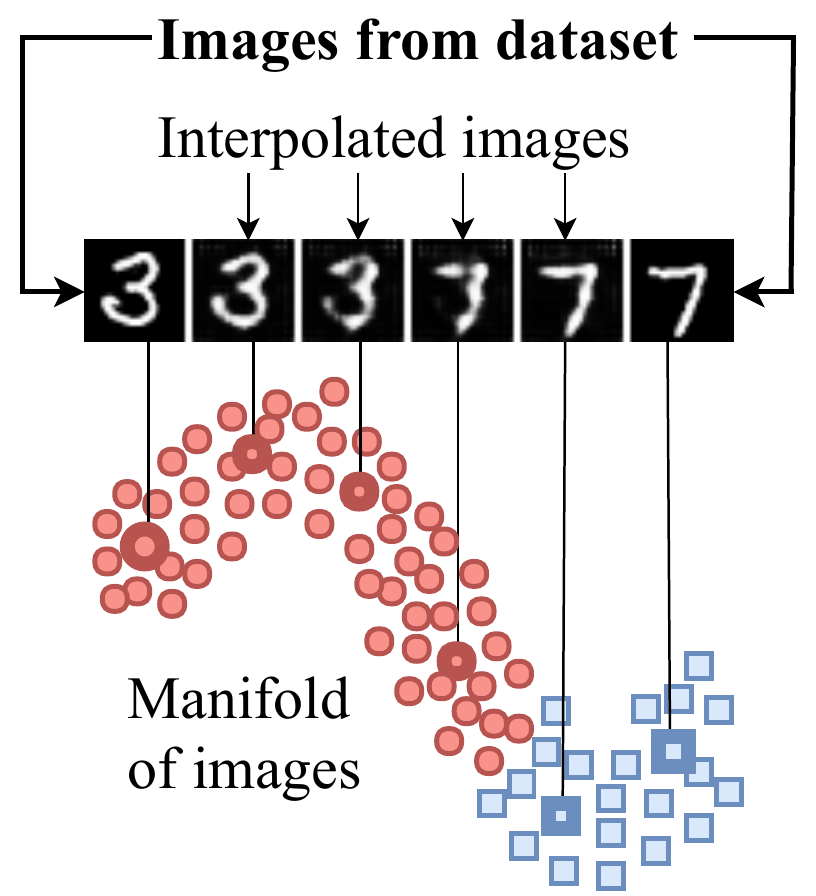}
\end{center}
   \caption{This figure shows how the latent representation of a GAN gives us a handle for the manipulation of the manifold. It shows how a GAN with an autoencoder generator allows us to navigate the manifold of images. Here, only the first and final images are among the natural dataset images, in this case a three and an seven. The in-between images have been obtained by interpolating in the latent space of the GAN, which allows us to effectively navigate the manifold.}
\label{fig:fig7}
\end{figure}

\begin{figure}[t!]
\begin{center}
   \includegraphics[width=1.0\linewidth]{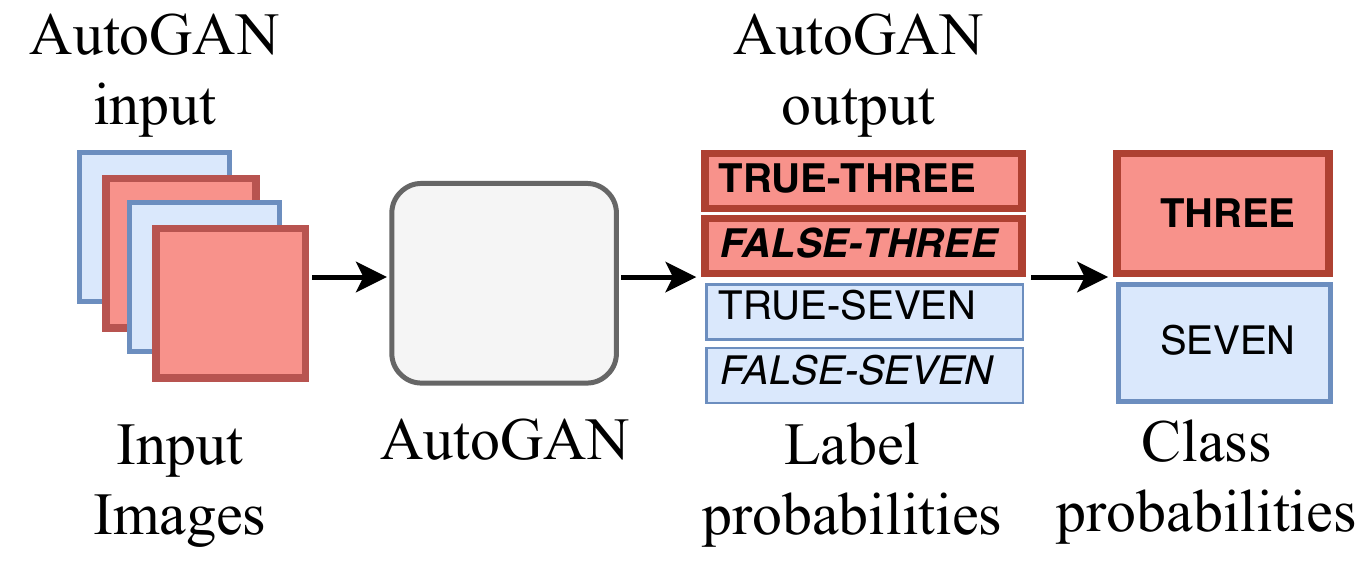}
\end{center}
   \caption{Illustration of how AutoGAN finds the label of an input image. The probability that an image is a digit three is the sum of the probability that the image is a TRUE-THREE and the probability that the image is a FALSE-THREE. The same for digit seven.}
\label{fig:fig4}
\end{figure}

\paragraph{GAN choice} We chose to use a GAN in our approach because GANs have been shown to be very good at reconstructing manifolds of natural image datasets in original high-dimensional input spaces. GANs also enable navigation and manipulation of the manifold. AutoGAN assumes no augmentation of training dataset with adversarial samples. For the purpose of repositioning the existing data points, a GAN~\cite{goodfellow2014generative, radford2015unsupervised} is well-suited since it is the best current method to reconstruct manifolds of natural image datasets in original high-dimensional input spaces. In particular, GANs where the generator is an autoencoder result in more realistic results~\cite{zhu2016generative} because they avoid images falling off the natural image manifold by not starting from random input vectors. The latent representation of the generator in such a GAN is essentially the low-dimensional manifold of the images and it allows us to navigate and manipulate the manifold of images, for example to do interpolation between two points, as illustrated also by Zhu \textit{et al.}~\cite{zhu2016generative} and described by Goodfellow \textit{et al.}~\cite{goodfellow2014generative}. The interpolation between two data points depicted in Figure~\ref{fig:fig7} cannot be done in the original input space, but can be achieved by training the dataset in a GAN that has an autoencoder generator. Then two data points are passed through the generator up to the latent space, where their latent representation is obtained. Other points between them are found using linear interpolation in the latent space. Finally, all the points are projected to the output of the generator. Thus, a smooth transition between the input images along the manifold is obtained, which is shown in Figure~\ref{fig:fig7}. AutoGAN incorporates class labels into both TRUE and FALSE labels that a normal GAN has. For a binary classification problem with digits three and seven, this means having 4 labels in the discriminator: TRUE-THREE, TRUE-SEVEN, FALSE-THREE, FALSE-SEVEN as depicted in Figure~\ref{fig:fig4}. The presence of labels that distinguish TRUE and FALSE images ensures that, just like in GANs, the manifold at the output of the autoencoder is aimed to be a reconstruction of the input manifold of natural images. The additional presence of class labels means that the discriminator is also aiming to get the classification right at the same time. This approach of using a GAN discriminator with more than two labels has been used for label efficiency~\cite{odena2016semi} as well as for improved image quality~\cite{salimans2016improved, odena2016conditional}. In contrast to these methods, AutoGAN has twice the number of original labels.

\begin{figure}[t!]
\begin{center}
   \includegraphics[width=1.0\linewidth]{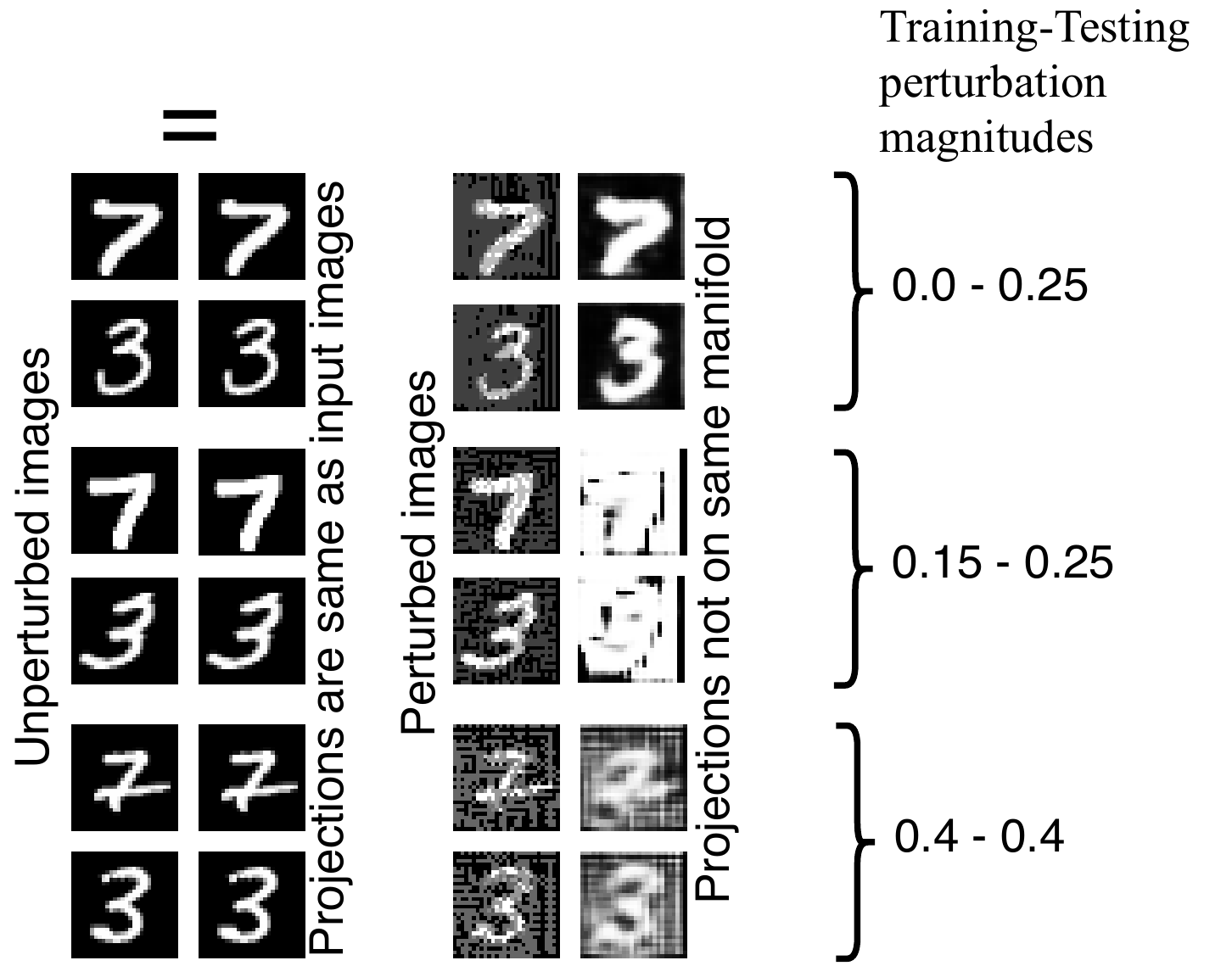}
\end{center}
   \caption{We show projections of original/unperturbed testing images from MNIST dataset and projections of the same images perturbed with the FGSM method at different magnitudes. The results are for 3 cases of training and testing perturbation magnitudes in AutoGAN. On the left, we see that original, unperturbed images get projected onto themselves, meaning that that they are projected onto the original input manifold. On the right are perturbed images of original MNIST dataset images and their projections onto the output manifold. The images on the right with the 0.15-0.25 perturbation magnitudes indicate that perturbed images are not always projected onto the original manifold -- their projections do not look like MNIST images. This indicates that AutoGAN can also enhance the manifold for better classification. The projections of perturbed images with the 0.15-0.25 perturbation magnitudes look like they have preserved only the edges of the digits and have disposed of the perturbations elsewhere.}
\label{fig:fig2}
\end{figure}

\paragraph{AutoGAN findings} We have discovered through our experiments in Section~\ref{autoganManifold}, illustrated in Figure~\ref{fig:fig2}, that AutoGAN projects unperturbed points onto themselves in the output manifold. In other words, AutoGAN does preserve the original manifold. However, we have also found cases where perturbed images do not seem to have been projected onto the original manifold -- images on the right of Figure~\ref{fig:fig2}. From these findings, we conclude that, in addition to preserving the original manifold, AutoGAN enhances it while also projecting points onto it for better classification. In the middle case on the right side of Figure~\ref{fig:fig2}, it seems that the projection of perturbed data points preserves the edges of the digits for better classification.

\begin{figure}[t!]
\begin{center}
   \includegraphics[width=1.0\linewidth]{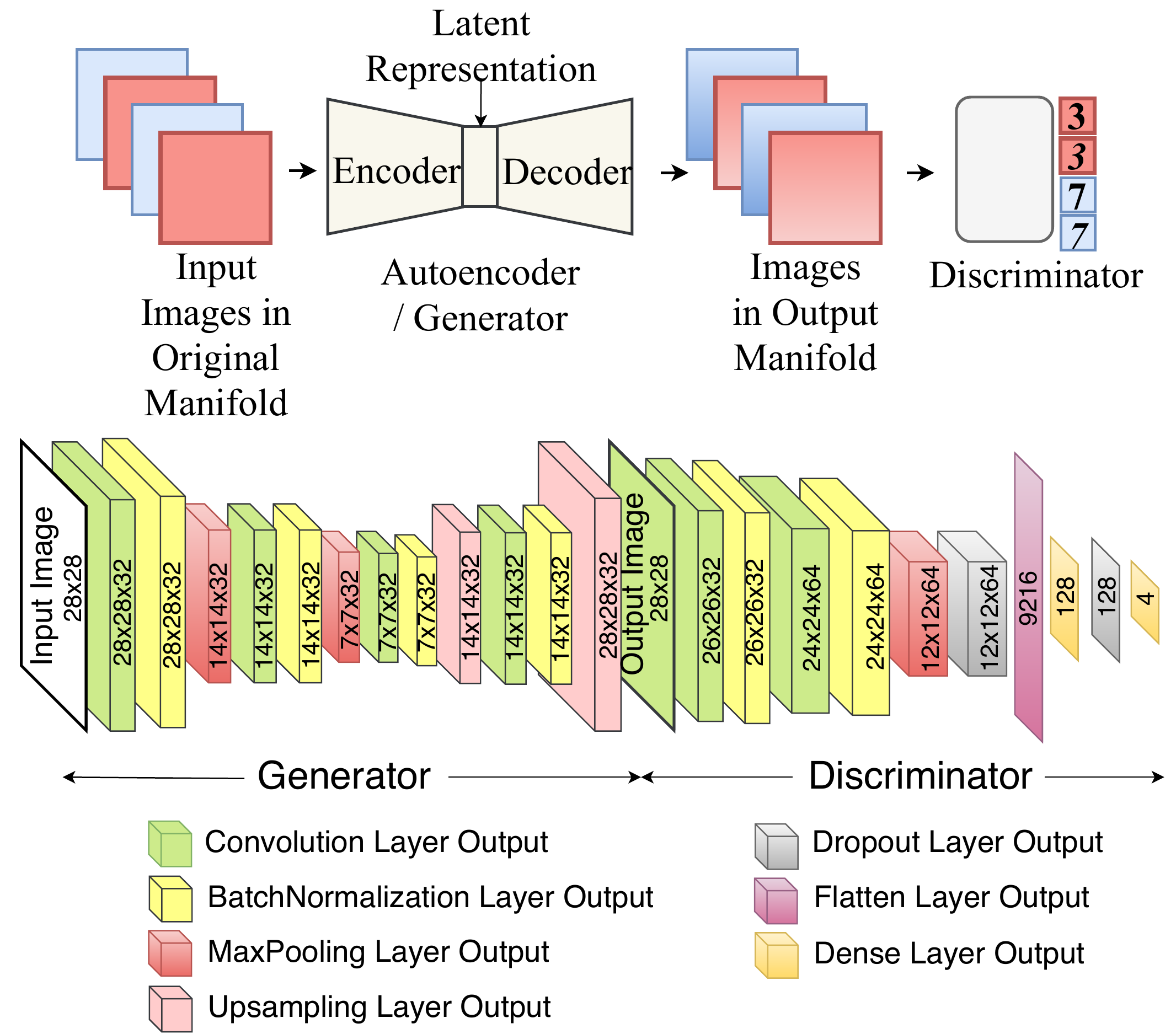}
\end{center}
   \caption{Illustration of AutoGAN architecture and the convolutional network architecture. The generator is an autoencoder where the size of the latent representation is smaller than the size of the input and output images. The discriminator outputs four labels, instead of the usual two labels, TRUE and FALSE, in GANs. We display also the layers of the generator and the discriminator.}
\label{fig:fig5}
\end{figure}

\paragraph{AutoGAN architecture} Figure~\ref{fig:fig5} shows that our design uses a GAN with an autoencoder generator. Additionally, in our design, the discriminator has four labels instead of two. In GANs, there are two labels, TRUE and FALSE to make the GAN reconstruct the manifold of images. In AutoGAN, we incorporate the image classes into the discriminator TRUE and FALSE labels. This is done with the purpose of reconstructing a manifold that leads to better classification.

The generator of the AutoGAN is an autoencoder that takes MNIST images as input and outputs images of the same size as MNIST images. The first part of the autoencoder has two convolutional layers with 32 feature maps each and ReLu activation. Each Convolutional layer is followed by a 2-by-2 MaxPooling layer. The second part of the autoencoder also has two convolutional layers with 32 feature maps each and ReLu activation. A 2-by-2 MaxPooling layer follows each Convolutional layer. It also has a final convolutional layer with just one feature map and sigmoid activation. All convolutional layers use 3-by-3 kernels. The autoencoder is compiled with the Adadelta optimizer and the loss is binary crossentropy. The layers of the generator are discplayed in Figure~\ref{fig:fig5}. The generator is compiled with the Adadelta compiler and the loss is binary crossentropy. Learning rate values are default values.

The discriminator is a neural network with two convolutional layers with 3-by-3 kernels and ReLu activations. The two convolutional layers have 32 and 64 feature maps and are followed by a 2-by-2 maxpooling layer, and then a dropout layer with a $0.25$ rate. This is followed by a dense layer with 128 units, another dropout layer with a $0.5$ rate, and finally a dense layer of four units. All layers except the final one have ReLu activation, while the final dense layer has softmax activation. The discriminator is compiled with the Adadelta compiler and the loss is categorical crossentropy. The default classifier that is used to compare the performace of AutoGAN against has the exact same architecture, with the difference that the final dense layer only has 2 units. The adversarial model is compiled with an SGD optimizer and with binary crossentropy loss. The layers of the discriminator are discplayed in Figure~\ref{fig:fig5}. The learning rates have default values.

\subsection{Classification with AutoGAN discriminator}

AutoGAN uses the discriminator for classification after training on the dataset by summing up label probabilities.
In a GAN, the discriminator uses the labels TRUE and FALSE to discriminate between original input and generated input. In AutoGAN, the discriminator incorporates the class distinction into its TRUE and FALSE labels so as to distinguish the digits three and seven. Within each of the original labels, there are two cases for each class. As a result, the AutoGAN discriminator has 4 labels: TRUE-THREE, FALSE-THREE, TRUE-SEVEN, FALSE-SEVEN. During testing, an image first goes through the generator, which projects it onto the output manifold. Then it is classified by the discriminator, yielding four probabilities, one for each label. The probability of each class -- digit three or seven -- is calculated as the sum of the probabilities that this image belongs to that class, either as original input or as a generated image. For example, the probability that an image is a digit three is the sum of the probability of label TRUE-THREE and the probability of label FALSE-THREE. The image gets classified to the class with the highest probability. This is illustrated in Figure~\ref{fig:fig4}.

\section{Experiments and Results}

For FGSM adversarial attacks, we assume that the attacker has access to the neural network model. For attacks with random Gaussian noise, we don't assume access to the neural network since the attacks are generated randomly.

We used the MNIST dataset~\cite{lecun1998mnist} for binary classification of digits three and seven, following Goodfellow \textit{et al.}~\cite{goodfellow6572explaining}. We implementated our approach with Keras~\cite{chollet2015keras}.

The training is done in random batches of 32 images. After 60 iterations, new testing samples and new training samples (when applicable) are generated. The best result within 50 epochs is reported. We report the mean of three measurements for all AutoGAN measurements and default classifier measurements. Because there was no variance in repeated FGSM measurements, we report single measurements for FGSM results. We benchmark AutoGAN robustness to FGSM adversarial attack using v2.1.0 of CleverHans~\cite{papernot2018cleverhans}. The training set gets modified by the FGSM method with a range of max norm $\epsilon$ values: 0.0, 0.1, 0.2, 0.3, 0.4. The testing set is modified by the FGSM method with a range of max norm $\epsilon$ values: 0.1, 0.2, 0.3, 0.4. We also benchmark AutoGAN robustness to attacks in random adversarial directions, where image pixels have Gaussian noise added with a $0$ mean and up to $1$ standard deviation. CleverHans~\cite{papernot2018cleverhans} takes as input the original MNIST dataset, the classifier (in our case, the AutoGAN discriminator), and the magnitude of perturbation and outputs adversarial samples. At each iteration, new training (if augmenting the training dataset) and testing get generaated. The testing samples are used to get intermediate accuracy results from the discriminator classifier. Another iteration of the training follows where the new adversarial training samples are used.

\subsection{AutoGAN outperforms FGSM in FGSM adversarial attacks} \label{fgsmExp}

For adversarial attacks in FGSM adversarial directions, we compare AutoGAN accuracy to FGSM~\cite{goodfellow6572explaining} method accuracy on adversarial samples generated with the FGSM method. We can categorize the cases when AutoGAN accuracy is distinctly better than FGSM method accuracy:

\begin{itemize}
  \item Figure~\ref{fig:fig0} shows results when the training dataset has not been augmented with adversarial samples -- the first column. This includes the best improvement AutoGAN makes, from 1\% with FGSM method to 89\% with AutoGAN.
  \item Figure~\ref{fig:fig0} also shows results when the magnitudes of perturbation in training and testing do not match -- bottom right and top left of the chart. Here, AutoGAN outperforms FGSM method by up to 25\% points.
\end{itemize}

In general, AutoGAN benefits from adversarial training, which can be seen by comparing AutoGAN accuracy values in the first column (where there is no augmentation) with AutoGAN accuracy values in the other columns (where there is augmentation).

\begin{figure}[t!]
\begin{center}
   \includegraphics[width=1.0\linewidth]{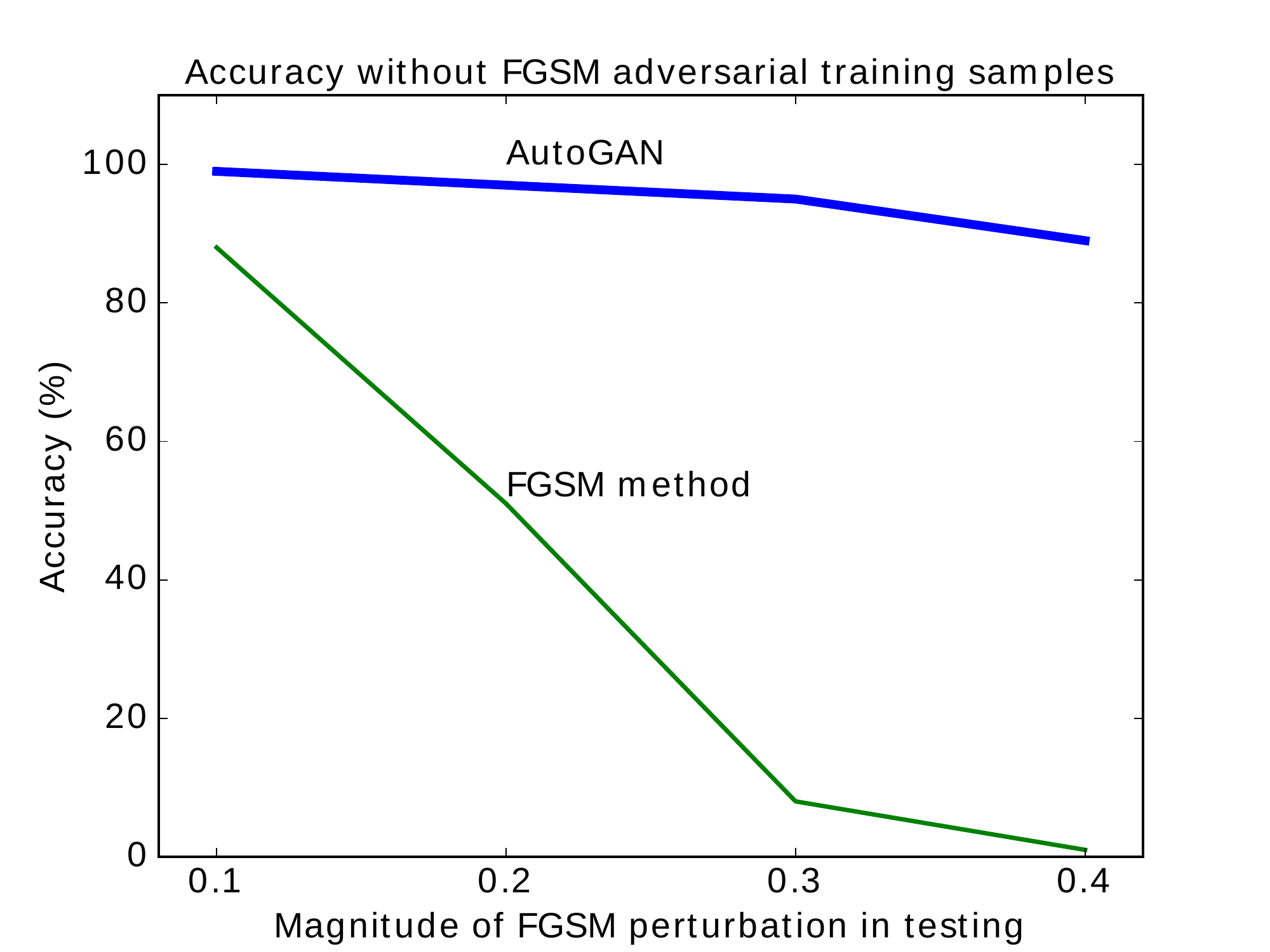}
\end{center}
   \caption{We compare the performance of AutoGAN and FGSM when trained without adversarial samples and tested with adversarial samples of various magnitudes of FGSM perturbation. Despite not being trained with adversarial samples, the AutoGAN method maintains a high accuracy when tested with adversarial samples of various magnitudes. On the other hand, the FGSM method and the default neural network classifer are able to handle only a small magnitude of FGSM perturbation; beyond that the accuracy values go down consistently with the magnitude of the FGSM perturbation.}
\label{fig:fig1}
\end{figure}

Figure ~\ref{fig:fig1} focuses on the first column of data in Figure~\ref{fig:fig0}, because this is where AutoGAN's improvement over FGSM method is the largest. This is the case where there is no adversarial training, the original training dataset has not been augmented with adversarial samples. This is also the likely scenario in machine learning applications, where the kind of attack is not known in advance. From Figure ~\ref{fig:fig1}, we can see that the AutoGAN maintains high accuracy even for high magnitudes of adversarial perturbation in testing. In the best case, the accuracy goes up to 87\% from FGSM's 1\%.

\subsection{AutoGAN outperforms default classifier on attacks in random adversarial direction} \label{gaussExp}

\begin{figure}[t!]
\begin{center}
   \includegraphics[width=1.0\linewidth]{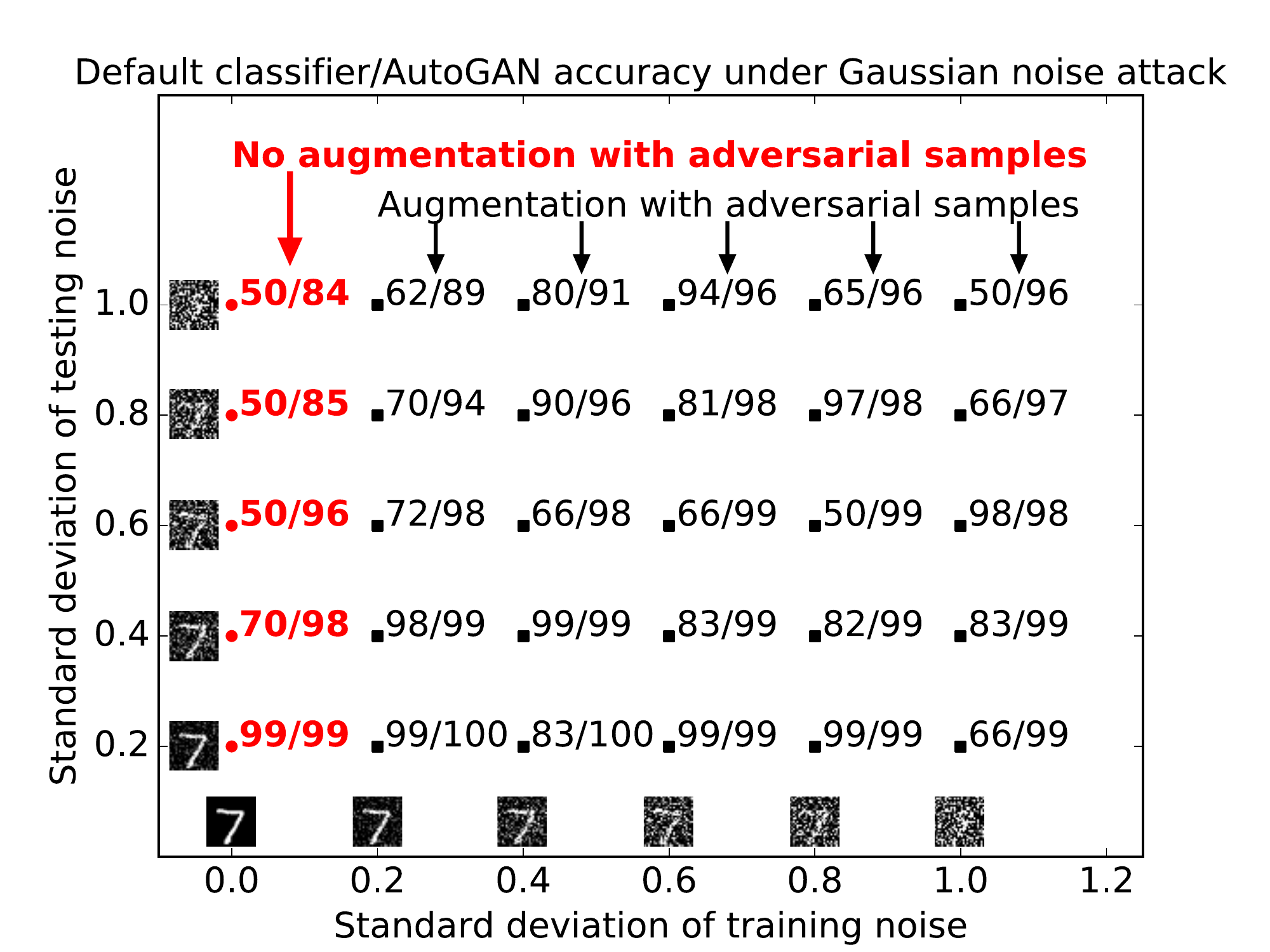}
\end{center}
   \caption{We compare the performance of a default neural network classifier and AutoGAN when trained with and without adversarial samples and tested with adversarial samples, both generated by adding Gaussian noise of various magnitudes. Here, we see that AutoGAN performance always exceeds default classifier performance, by up to 49\% points. In particular, the first column where there is no augmentation of the training dataset with adversarial samples displays the most benefit from AutoGAN. The structure of the default classifier is the same as that of the AutoGAN discriminator, except the number of labels is two, one for each class. All results are means of three measurements.}
\label{fig:fig3}
\end{figure}

To investigate whether AutoGAN is capable of handling other kinds of adversarial perturbation as well, we test it on samples with perturbation in random directions in the input space -- Gaussian noise added independently to each pixel with a $0$ mean and standard deviation up to $1$. Figure~\ref{fig:fig3} shows the accuracy results for AutoGAN in comparison to a default classifier that has the same architecture as the AutoGAN discriminator but only two labels, one for each class instead of the four labels the discriminaotr has. All results are means of three measurements. The results on Figure~\ref{fig:fig3} show that AutoGAN accuracy exceeds the accuracy of the default classifier, by up to 49\% points. AutoGAN performs better than a default classifier in both cases with or without augmentation of the training dataset with adversarial samples.

\subsection{AutoGAN enhances the manifold} \label{autoganManifold}

To investigate how the shape of the output manifold changes in comparison to the input manifold, we project images onto the output manifold. To project the images, we pass them through the autoencoder generator. The experiments are performed for several values of perturbation magnitudes in training and testing and the results of the experiments are displayed in Figure~\ref{fig:fig2}. On the left are unperturbed images and their projections, where we can see that the projections are the same as the original images. This indicates that AutoGAN does preserve the original manifold. On the right are perturbed images and their projections, where we can see that the projections are different from the input perturbed images. More importantly, we can see in the middle digits on the right, the projections do not resemble digits from the MNIST dataset. This indicates that AutoGAN has not projected these perturbed images on the original manifold. From this we conclude that AutoGAN must have enhanced the original manifold and then projected perturbed images onto the enhancement.

\subsection{AutoGAN generator and discriminator both needed against adversarial attacks} \label{bothNeeded}

We further investigated whether the reshaping of the manifold by the AutoGAN generator was enough for better classification or whether the discriminator was needed as well for better classification. In other words, could the discriminator be replaced with a default classifier? We evaluated a default classifier without augmenting the training dataset with adversarial samples. We trained AutoGAN as previously described. Then we trained a default classifier on data points after they go through the generator. The default classifier had the same structure as the discriminator with the only difference being the number of labels -- the default classifier had only two class labels. As Table \ref{table:results} shows, the classification results were not good when replacing the AutoGAN discriminator with a default classifier. As a result, we conclude that the classification logic of the AutoGAN is contained in both the generator and the discriminator. They are both needed in achieving the AutoGAN accuracy results. This has security implications because if the logic for countering adversarial attacks is contained in both the generator and the discriminator, it means that AutoGAN is more robust against attackers since only the discriminator can be exposed to attackers.

\begin{table}[]
\caption{Classification results after using a default classifier instead of the AutoGAN discriminator for the classification}
\centering
\begin{tabular}{| l | l | l | l | l |}
\hline
Magnitude of FGSM & & & & \\
testing perturbation & 0.1 & 0.2 & 0.3 & 0.4 \\

\hline
Accuracy (\%) & 50 & 50 & 1 & 3 \\
\hline
\end{tabular}
\label{table:results}
\end{table}

\section{Conclusions}

We have designed, implemented and evaluated a novel AutoGAN classifier that is resilient against adversarial attacks. Unlike other defense methods, AutoGAN employs at the same time two defenses, which we think explains its resilience -- AutoGAN changes the model to make it more robust and also moves adversarial points to make them less adversarial. Though AutoGAN can benefit from augmenting its training dataset with adversarial samples, it maintains accuracy even without them. This matters since attack type might not be known and adversarial samples might not be available. AutoGAN achieves 89\% accuracy compared to FGSM's 1\% in the best case without augmentation. AutoGAN is also able to maintain high accuracy for different magnitudes of perturbation in training and testing, exceeding FGSM accuracy by up to 25\% points on adversarial samples perturbed by the FGSM method. Against adversarial attacks in random adversarial directions, AutoGAN performs consistently well, even without augmentation of the training dataset.

\section{Acknowledgements}
This research was sponsored by the U.S. Army Research Laboratory and
was accomplished under Cooperative Agreement Number W911NF-13-2-0045
(ARL Cyber Security CRA). The views and conclusions contained in
this document are those of the authors and should not be interpreted
as representing the official policies, either expressed or implied,
of the Army Research Laboratory or the U.S. Government. The U.S.
Government is authorized to reproduce and distribute reprints for
Government purposes notwithstanding any copyright notation here on.

{\small
\bibliographystyle{plain}
\bibliography{egbib}
}

\end{document}